\documentclass[runningheads,a4paper]{llncs}

\usepackage{amssymb}
\usepackage{graphicx}
\usepackage{url}
\usepackage{enumerate}
\usepackage{cleveref}
\usepackage[caption=false]{subfig}
\usepackage{todonotes}
\usepackage{enumitem}
\usepackage{comment}

\urldef{\mailsa}\path|{f.delvigna, m.petrocchi, m.tesconi}@iit.cnr.it|
\urldef{\mailsb}\path|{alessandro.tommasi, cesare.zavattari}@lucense.it|

\newcommand{\keywords}[1]{\par\addvspace\baselineskip
\noindent\keywordname\enspace\ignorespaces#1}

\begin{document}

\mainmatter  

\title{Semi-supervised knowledge extraction\\
for detection of drugs and their effects
}

\titlerunning{Semi-supervised knowledge extraction for detection of drugs and their effects}

%
%
\author{Fabio Del Vigna\inst{1,2} \and Marinella Petrocchi\inst{2} \and Alessandro Tommasi\inst{3} \and \\Cesare Zavattari\inst{3} \and Maurizio Tesconi\inst{2}}
\authorrunning{F.~Del~Vigna {\it et al.}}

\institute{Dept. of Information Engineering, University of Pisa, Italy
\and
Institute of Informatics and Telematics (IIT-CNR), Pisa, Italy\\
\and
LUCENSE SCaRL, Lucca, Italy\\
\mailsa\\
\mailsb}

%
%

\toctitle{Semi-supervised knowledge extraction for detection of drugs and their effects}
\tocauthor{Fabio Del Vigna}
\maketitle

\begin{abstract}
New Psychoactive Substances (NPS) are drugs that lay in a grey area of legislation, since they are not internationally and  
officially banned, possibly leading to their not prosecutable trade. The exacerbation of the phenomenon is that NPS can be easily sold and bought online.  
Here, we consider large corpora of textual posts, published on online forums specialized on drug discussions, plus a small set of known substances and associated effects, which we call seeds. 
We propose a semi-supervised approach to knowledge extraction, applied to the detection of drugs (comprising NPS) and effects from the corpora under investigation. Based on the very small set of initial seeds, 
the work highlights how a contrastive approach and context deduction are effective in detecting substances and effects from the corpora. 
Our promising results, which feature a F1 score close to 0.9, pave the way for shortening the detection time of new psychoactive substances, once these are discussed and advertised on the Internet.
\keywords{Text mining, NPS detection, NPS data mining, drugs forums, social media analysis, machine learning, automatic classification.}
\end{abstract}

\section{Introduction}
\label{sec:intro}
US and European countries are facing a raising emergency: the trade of substances that lay in a grey area of legislation, known as New Psychoactive Substances (NPS). The risks connected to this phenomenon are high: every year, hundreds of consumers get overdoses of these chemical substances and hospitals have difficulties to provide effective countermeasures, given the unknown nature of NPS. 
Government and health departments are struggling to monitor the market to tackle NPS diffusion, forbid NPS trade and sensitise people to the harmful effects of these drugs\footnote{\url{http://www.emcdda.europa.eu/start/2016/drug-markets\#pane2/4 };  All URLs in the paper have been accessed on July 10, 2016.}. Unfortunately, legislation is typically some steps back and newer NPS quickly replace old generation of substances.
Also, the abuse of certain prescription drugs, like opioids, central nervous system depressants, and stimulants, is a widespread as an alarming trend, which can lead to a variety of adverse health effects, including addiction\footnote{\url{https://www.drugabuse.gov/publications/research-reports/prescription-drugs/director}}. 

The described phenomena are being exacerbating by the fact that online shops and marketplaces convey NPS through the Internet~\cite{Schmidt2011}. Moreover, specialised forums offer a fertile stage for questionable organisations to promote NPS, as a replacement of well known drugs. 
Forums are contact points for people willing to experiment with new substances or looking for alternatives to some chemicals.

In this work, we consider the myriads of posts published 
on two big drugs forums, namely Bluelight\footnote{\url{http://www.bluelight.org}} and Drugsforum\footnote{\url{https://drugs-forum.com}}. 
Posts consist of natural language, unstructured text, which, generally speaking, can be analysed with text mining techniques to discover meaningful information, useful for some particular purposes~\cite{Witten2004}. 
%
%
%
We propose DAGON (DAta Generated jargON), a novel, semi-supervised knowledge extraction methodology, and we apply it to the posts of the drugs forums, with the main goals of: i) detecting substances and  their effects;
ii) put the basis for linking each substance to its effects. A successful application of our technique is paramount: first, we envisage the possibility to shorten the detection time of NPS; then, it will be possible to group together different names that refer to the same substance, as well as to distinguish between different substances, commonly referred to with the same name (such as ``Spice"~\cite{Schifano2009}) and timely detect changes in drug composition over time~\cite{Davies2010}. Finally, knowing the effects tied to novel substances, first-aid facilities may overcome the current 
difficulties to provide effective countermeasures.

While traditional supervised techniques usually require large amount of hand-labeled data, our proposal features a semi-supervised learning approach in order to minimize the work required to build an effective detection system.
Semi-supervised learning exploits unlabeled data to mitigate the effect of insufficient labeled data on the classifier accuracy. This specific approach attempts to automatically generate high-quality training data from an unlabeled corpus.
With very little information, our solution is able to achieve excellent detection results on drugs and their effects, with an FMeasure close to 0.9.

The paper is structured as follows. 
The next section describes our data sources. In Section~\ref{sec:semi-super}, we introduce our semi-supervised methodology. Section~\ref{sec:exp} presents a set of experiments and results. Section~\ref{sec:relwork} provides related work on mining drugs over the Internet and it discusses text analysis approaches, highlighting differences and similarities with our proposal. Finally, Section~\ref{sec:conc} concludes the paper. 

\section{Datasets}
\label{sec:datasources}
The approach in this work is tested over two different large data sources, in order to consider a variety of contents and information, and to push the automatic detection of drugs. We collected more than a decade of posts from  Bluelight and Drugsforum. As shown in Table~\ref{tab:forums}, the available data comprises more than half million users and more than 4.6 million posts. Data was collected through web scraping and stored in a relational database for further querying. These forums were early and partially analysed in~\cite{Soussan2014} and then explored in detail~\cite{Delvigna2016}. Here, we present the very same datasets to show how it is possible to extract knowledge from text using few seeds as the starting point for the algorithm introduced in Section~\ref{sec:semi-super}.

\begin{table}[ht]
    \centering
    \begin{tabular}{|c|c|c|c|c|}
        \hline
        \bf{Forum} & \bf{First post} & \bf{Last post} & \bf{Tot posts} & \bf{Users}\\ 
        \hline
        Bluelight & 22-10-1999 & 09-02-2016 & 3,535,378 & 347,457\\
        Drugsforum & 14-01-2003 & 26-12-2015 & 1,174,759 & 220,071\\
        \hline
    \end{tabular}
    \caption{Drug forums: Posts and Users}
    \label{tab:forums}
\end{table}

\vspace{-15mm} 
\subsection{Seeds}
\label{sec:seeds}
We have downloaded a list of 416 drug names of popular psychoactive substances, including the slang which is adopted among consumers to commonly name them, from the website of the project \emph{Talk to Frank}\footnote{\url{http://www.talktofrank.com}} and a dataset containing 8206 pharmaceutical drugs retrieved from Drugbank\footnote{\url{http://www.drugbank.ca}}. This list constitutes a ground truth for known drugs.

Also, we collected a list of 129 symptoms that are typically associated to substance assumption.

\section{The DAGON methodology (DAta Generated jargON)}
\label{sec:semi-super}
In this section, we introduce DAGON, a methodology that will be applied in Section~\ref{sec:exp} for the 
task of identifying new ``street names'' for drugs and their effects. A street name is the name a substance is usually referred to amongst users and pushers. 

The task of name identification can be split into two subtasks:
\begin{enumerate}[label=(\alph*)]
\item \label{lab:ident}Identifying text chunks in the forums, which represent candidate drug names (and candidate drug effects); 
\item Classifying those chunks as drugs, effects, or none of the above.
\end{enumerate}
The first subtask - identification of candidates - could be tackled with different approaches, including a noun-phrase identifier\footnote{A noun-phrase is a phrase that plays the role of a noun such as ``the kid that Santa Claus forgot".}, usually based on a simple part-of-speech-based grammar, or on a technique akin to the identification of named entities, as in~\cite{M98-1002}.

In this work, the identification of candidates is based on domain terminology extraction techniques based on a contrastive approach similar to~\cite{Penas01corpus-basedterminology}. Essentially, we identify chunks of texts that appear to be especially significant in the context of drug forums. Based on the frequency in which terms appear both in the posts of drugs forums and in contrastive datasets dealing with different topics, we extract the most relevant terms for the forums. We have extracted unigrams, 2-grams, and 3-grams. This approach does not require English specific annotated resources and, thus, it can scale easily to different languages. 

The second subtask is a classification problem. Following a supervised approach would have required to have annotated posts and use them as the training set for our classifier. Instead, we have chosen to work on unlabeled data (i.e., the posts on the drugs forums, see Section~\ref{sec:datasources}) and to exploit the external list of seeds introduced in Section~\ref{sec:seeds}.

We represent a candidate by means of the words found along with it when it was used in a post, selecting windows of $N$ characters surrounding the candidate whenever it was used in the dataset. 
Hereafter, we call {\it context} (of a candidate) the text surrounding the term of interest.

Thus,  we have shifted the problem: from classifying candidate street names to the classification of their contexts, which are automatically extracted from the unlabeled forum datasets.

It is worth noting that, in the drugs scenario, there would be at least 3 classes, i.e.,  Substance, Effect, and ``none of the above'' - the latter to account for the cases where the candidate does not represent substances and effects.
However, the seed list at our disposal consists of flat lists of substances/effects names, provided with no additional information (Section~\ref{sec:seeds}). 
Therefore, in the following, we will first automatically identify positive examples for the two classes (Substance and Effect), training a classifier on them, and then we will tune the classifier settings to determine when a candidate does not fall in either.

Summarising, we have split the task of classifying a candidate into the following sub-tasks:
\begin{enumerate}[label=(\alph*)]
\item Fetch a set of occurrences of the term along with the surrounding text (forming in such a way the so called contexts). 
\item \label{item:classification}Classify each context along the 2 known classes (Section~\ref{sec:classification}). 
\item Determine a classification for the term given the classification result for the context related to that term (obtained at step \ref{item:classification}). 
%
\end{enumerate}
The single context classification task~\cite{Attardi98theseus:categorization} falls within the realm of standard text categorization, for which there is a rich literature.

\def\mcnt{\ensuremath{M^{ctx}}}
\def\mtrm{\ensuremath{M^{trm}}}
\def\tconf{\ensuremath{\theta_{p}}}
\def\tratio{\ensuremath{\theta_{c}}}

Hereafter, we detail the training phase for our classifier (\ref{sec:algorithm}), we give detail on the choice of seeds (\ref{sec:seedchoice}), we specify the procedure for classifying a new candidate (\ref{sec:classification}), and we illustrate a simple approach to link substances to their effects (\ref{sec:link}). 

\subsection{Training phase}
\label{sec:algorithm}
We are equipped with a list of examples for both the drugs and the effects, as described in Section~\ref{sec:seeds}. This list of entry terms is the training set for the classification task and we call it \textit{list of seeds}.

Each post in the target drug forums was indexed by a full-text indexer (Apache Lucene\footnote{\url{http://lucene.apache.org/}}) as a single document. 

The training phase is as follows:
\begin{enumerate}[label=(\roman*)]
\item Let $T_S$ and $T_E$ be the set of example contexts, for the Substance and Effects classes respectively, initialized empty. 
\item \label{lab:pickseed} From the lists of seeds, we pick a new seed (a drug name) for the Substance class and one (an effect name) for the Effects class. A seed is therefore an example of the corresponding class taken from the seed list (Section \ref{sec:seeds}). See Section~\ref{sec:seedchoice} for the heuristic to select a seed out of the list.
\item \label{lab:searchcontext} We use the full-text index to retrieve $M$ posts containing the seed $s$; we only use the bit of text surrounding the seed. In Section~\ref{sec:exp}, we will show how results change by varying $M$. We pick a window of 50 characters surrounding the searched seed.
\item \label{lab:stripcontext} We strip $s$ from the text, replacing it always with the same unlikely string (such as ``CTHULHUFHTAGN''), in order to avoid the bias carried by the term itself, but maintaining the position of the term in the phrase for classification purposes. We call the texts thus obtained $ctx_s$ (context of seed $s$).
\item We add the texts thus generated to the set of training examples for the category $C$ the seed belongs to (either $T_S$ or $T_E$)
\item We use the training examples to train a multiclass classification model $\mcnt$, which can be any multiclass model, as long as it features a measure (e.g., a probability) interpretable as a confidence score of the classification. In section \ref{sec:exp} we will show results when using SVM with linear kernel \cite{Chang01libsvm:a}.
\end{enumerate}

At the end of these steps, we have obtained a classifier of contexts ($\mcnt$), but as seeds (not contexts) are labeled, we are unable to assess its performance directly. We therefore define a classifier of candidate terms ($\mtrm$) using the method described later in Section~\ref{sec:classification}, the performance of which we can assess against the seed list. This allows us to optionally iterate back to step \ref{lab:pickseed}, in order to provide additional seeds to extend the training sets, and improve performances.

\begin{figure}[!ht]
    \centering
    \includegraphics[width=0.8\textwidth]{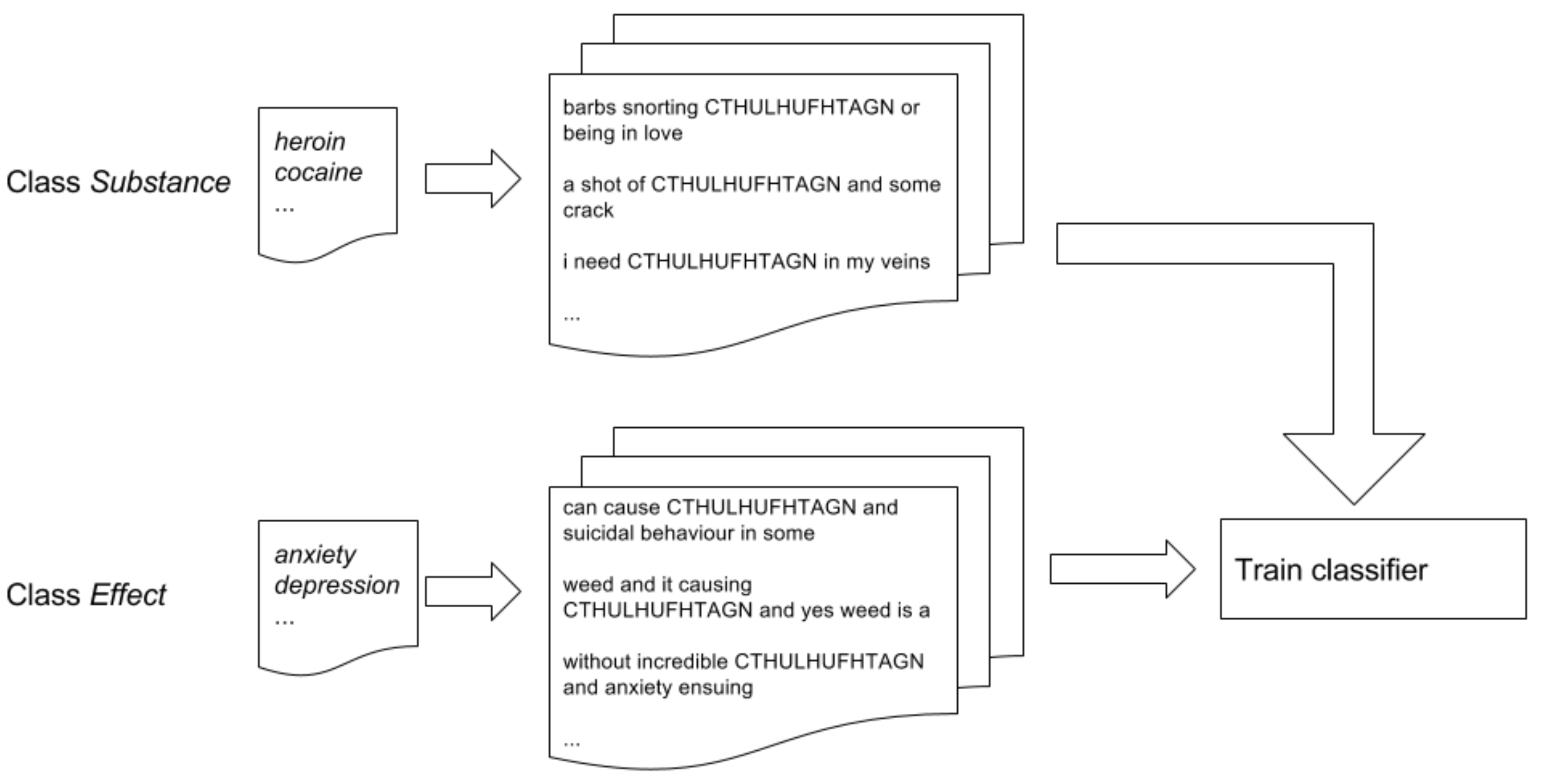}
    \caption{Training phase}\label{fig:fitting}
\end{figure}

The rationale behind this process is that drug (and effects) mentions will likely share at least part of their immediate contexts. Clearly, when a very small number of seeds is provided (e.g., 1 per class) there will be a strong bias in the examples ultimately used for training, which means that the resulting model will be overly specific to the type of drug used in the training. By providing more seeds, and with enough variety, the model will eventually become more generic to encompass the various drug types, and the relative differences in the contexts in which they are mentioned in the dataset.

\subsection{Choosing a seed}
\label{sec:seedchoice}
Obtaining a large seed list is often costly, since it may require to manually annotate texts, or to provide to the algorithm a initial set of words. Thus it is important to design a system with high performances that uses the minimum amount possible of seeds for the train phase.
Choosing an effective seed is paramount, and, in doing so, there are various aspects to consider:
\begin{enumerate}[label=(\alph*)]
\item \label{lab:frequentseed} Is the seed mentioned verbatim enough times in the data collection? Failing this, the seed will only serve to collect a small number of additional training elements, and it will not impact the model enough;
\item \label{lab:informativeseed} Is the seed adding new information? The most effective seeds are those whose contexts are misclassified by the current iteration of the classification model. In order to pick the most useful one, we could select, from the list of available unused seeds, those whose contexts are frequently misclassified. 
Using these seeds, the model is modified to address a larger number of potential errors.
\end{enumerate}
In information retrieval, Inverse Document Frequency~\cite{Salton:1988:TAA:54259.54260} (idf) is often used along with term frequency (tf) as a measure of relevance of a term, capturing the fact that a term is frequent, but not so frequent to be essentially meaningless (non-meaning words, such as articles and conjunctions, are normally the most frequent ones). A common way to address point \ref{lab:frequentseed} would therefore be using a standard tf$\cdot$idf metric. However, because our seeds list is guaranteed to only contain meaningful entries, we can safely select the terms occurring in more documents first (i.e., with an increasing idf).
We leave point \ref{lab:informativeseed} for future work.

\subsection{Classification of a new candidate}
\label{sec:classification}
At the end of the training phase, the classifier $\mcnt$ has been trained - on contexts of the selected seeds - to classify as either pertaining to substances or effects. 
Here, we describe the procedure by which, given a new candidate $c$, we establish what class (Substance or Effect) it belongs to.
The new candidates are chosen from the terms which are more relevant for the forums. Such terms are extracted according to the contrastive approach described in Section~\ref{sec:semi-super}, subtask \ref{lab:ident}.

The training phase produces a model $\mcnt$ by which contexts in which the term appears are classified -- we define here a model $\mtrm$ by which the term itself is classified into either Substance, Effect, or ``none-of-the-above''. $\mtrm$ is defined as a function of a candidate $c$ and the existing model $\mcnt$ as follows:
\begin{enumerate}
\item We apply steps \ref{lab:searchcontext} and \ref{lab:stripcontext} of the algorithm described in \ref{sec:algorithm} to obtain the contexts for $c$ ($ctx_c$).
\item We classify the elements of $ctx_c$ using $\mcnt$. We discard all categorizations whose confidence, according to the model, falls below a threshold \tconf, which we have experimentally set to 0.8 as a reference value.
\item We consider the remaining categorizations thus obtained. If a sizeable portion of them (\tratio, initially set to 0.6, we will show how results vary along with its value) belongs to the same class $C$, then $c$ belongs to $C$; otherwise it is left unassigned.
\end{enumerate}

\begin{figure}[!ht]
    \centering
    \includegraphics[width=0.7\textwidth]{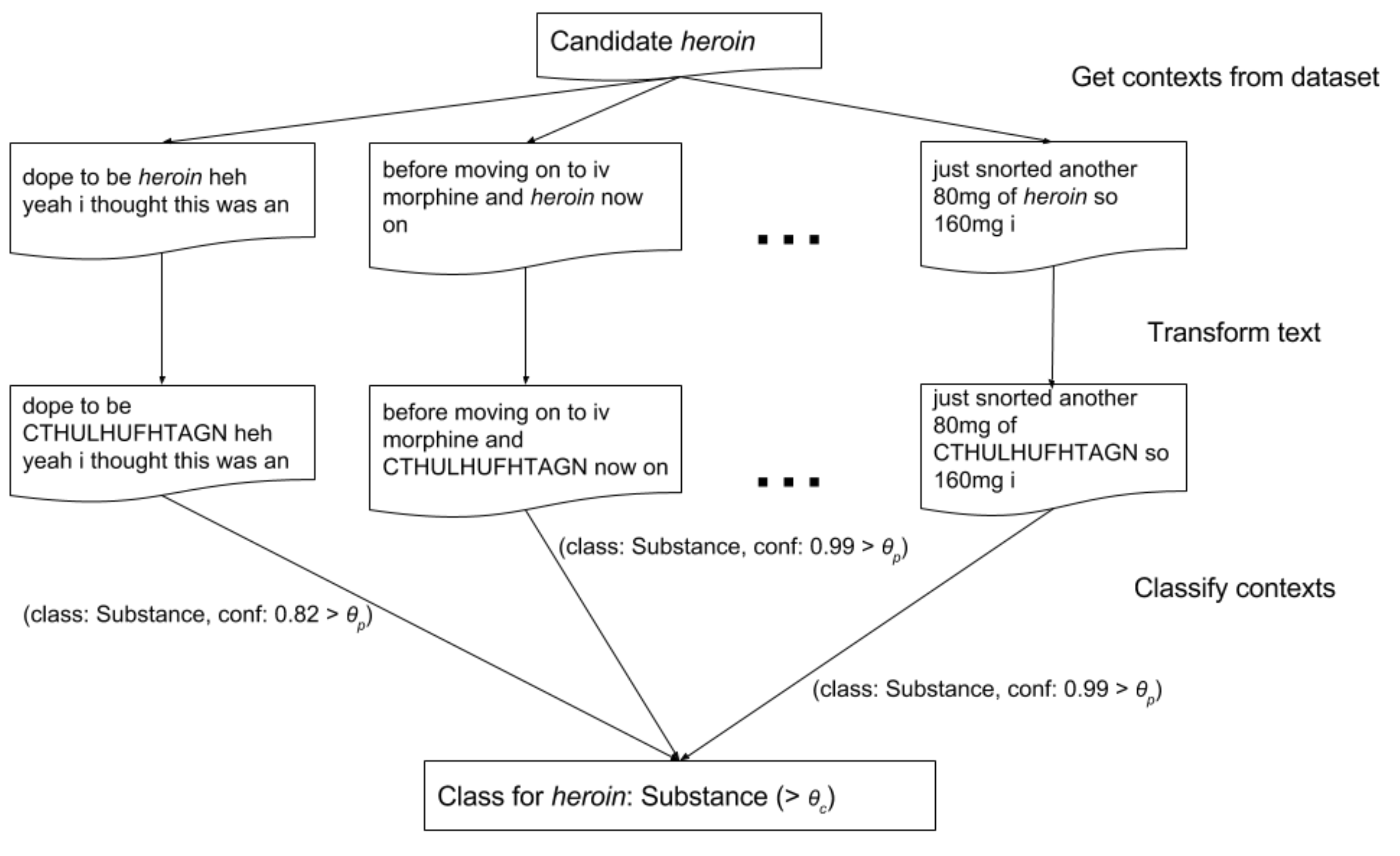}
    \caption{Classification of a new term}\label{fig:classify}
\end{figure}

In Figure~\ref{fig:classify} we give a high level graphical description of this process.

\subsection{Linking substances to effects}
\label{sec:link}
We outline here a simple procedure by which we can associate the substances mentioned in the drugs forums to the effects they produce.

When indexing a post, the significant terminology elements found in the post are linked to it as metadata. As introduced, the terminology elements have been extracted following a contrastive approach, as in~\cite{Penas01corpus-basedterminology}. 

We assume to have already tagged the terminology elements found in each post as referring to substances or effects, using the method described in Section~\ref{sec:classification}. 
Thus, when searching for mentions of a particular substance, we can correspondingly fetch, for each post the substance mention is found in, the relative metadata.
Then, from the matadata, we can sort the list of effects by frequency -- it is very likely that those effects are related to the searched substance.



As a simple example, let's suppose to have a single post, with 
Text: {\it heroin gave me a terrible headache}; 
Substances: [heroin]; 
Effects: [headache]. 

Intuitively, we can assume that [headache] is an effect of [heroin]. If we consider all the posts in our datasets where the substance [heroin] is among the metadata, and we count the most frequent metadata effects associated to [heroin], we can have an indication of the links between substances and effects. However many substances may appear in the same text. Thus, it is necessary to filter out the rarest links substance-effect since they are often due by chance. 
Section~\ref{sec:exp} will report on some findings we were able to achieve for our datasets about drugs and their effects. 

\section{Experiments}
\label{sec:exp}
We show a set of experiments on the data described in Section \ref{sec:datasources}.
%
%
%
First, from all the posts, we need to identify a list of candidates (unless we want to try and classify every term -- a possible, but undesirable strategy, to pinpont substances or effects out of which. Candidates are selected using a contrastive terminology extraction~\cite{Penas01corpus-basedterminology}, to identify terms and phrases common within the community and yet specific to it; this is the first subtask outlined in section \ref{sec:semi-super}. Then, we apply the $\mtrm$ classifier, described in Section \ref{sec:classification}, to assign to candidates either the class Substance or Effect or none of the above, and evaluate the performance of the classification. The intermediate $\mcnt$ classifier was trained using SVM with linear kernel \cite{Chang01libsvm:a}.

We report experiments and results for the Bluelight forum. The lists used to select seeds and to validate results have been described in Section~\ref{sec:seeds}. These lists represent 2 classes: Substance and Effect. 

It is worth noting that, for our experiments, we consider the intersection between the lists of seeds and the extracted terminology. This is necessary because:
i) items that are present in the lists may not be present in the downloaded dataset; ii) many terminological entries might be neither drug names nor drug effects.
The intersection contains 226 substances and 89 effects. Some of these will be used as seeds, the rest of the entries to validate the results.

 
 %
 The results are given in terms of three standard metrics in text categorization, based on true positives (TP - items classified in category $C$, actually belonging to $C$), false positives (FP - items classified in $C$, actually not belonging to $C$) and false negatives (FN - items not classified in $C$, actually belonging to $C$), computed over the decisions taken by the classifier: precision\footnote{$precision=\frac{TP}{TP+FP}$}, recall\footnote{$recall=\frac{TP}{TP+FN}$} and F1-micro averaged\footnote{harmonic mean of $precision$ and $recall$: $F1=2\cdot\frac{precision \cdot recall}{precision+recall}$}.

\begin{figure}[!ht]
    \centering
    \includegraphics[width=0.7\textwidth]{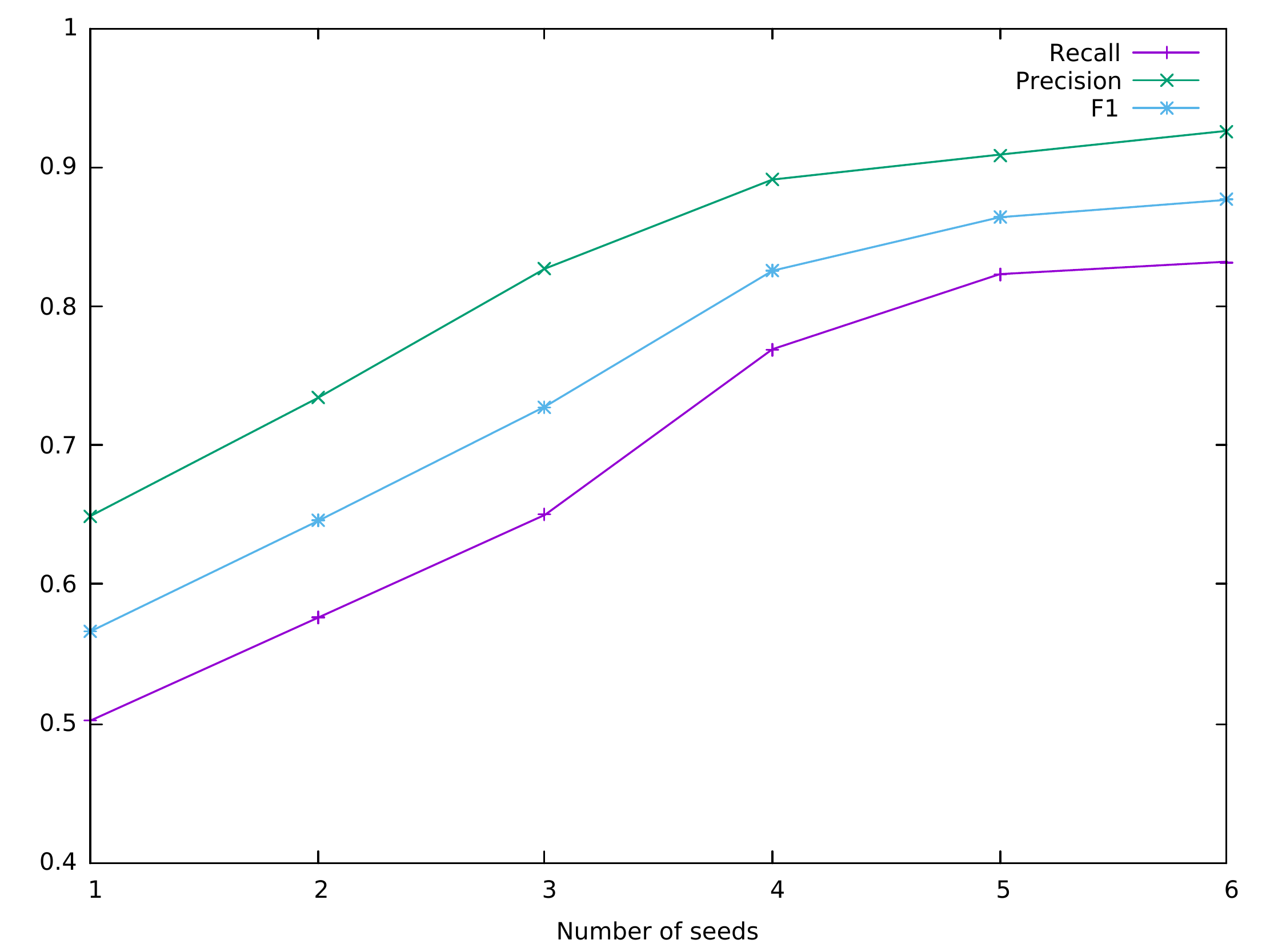}
    \caption{Recall, precision and F1 varying the number of seeds}
    \label{fig:fscore}
\end{figure}

 The first results are in Table~\ref{tab:result2classes} and Figure~\ref{fig:fscore}. Even though the training set is limited to a small number of entries, the results are interesting: with only 6 seeds, the proposed methodology  achieves a F1 score close to 0.88 (on the 2 classes - Substance and Effect). 
 With the aim of monitoring the diffusion of new substances, the result is quite promising, since it is able to detect unknown substances without human supervision. 
\begin{table}[ht]
    \centering
    \begin{tabular}{|c|c|c|c|}
        \hline
        \bf{\# of seeds} & \bf{Recall} & \bf{Precision} & \bf{F1}\\ 
        \hline
        1 & 0.502 & 0.649 & 0.566\\
        2 & 0.576 & 0.734 & 0.645\\
        3 & 0.65 & 0.827 & 0.728\\
        4 & 0.769 & 0.891 & 0.826\\
        5 & 0.823 & 0.909 & 0.864\\
        6 & 0.832 & 0.926 & 0.876\\
        \hline
    \end{tabular}
    \caption{Classification results for substances and effects, varying the number of seeds}
    \label{tab:result2classes}
\end{table}

{\bf Dealing with ``the rest''.} Finding mentions of new substances or effects means classifying candidates terms in either one class. Playing with thresholds, we can discard some candidates, as belonging to none of the two classes (see Section~\ref{sec:classification}). 

\begin{figure}[!ht]
    \centering
    \includegraphics[width=0.7\textwidth]{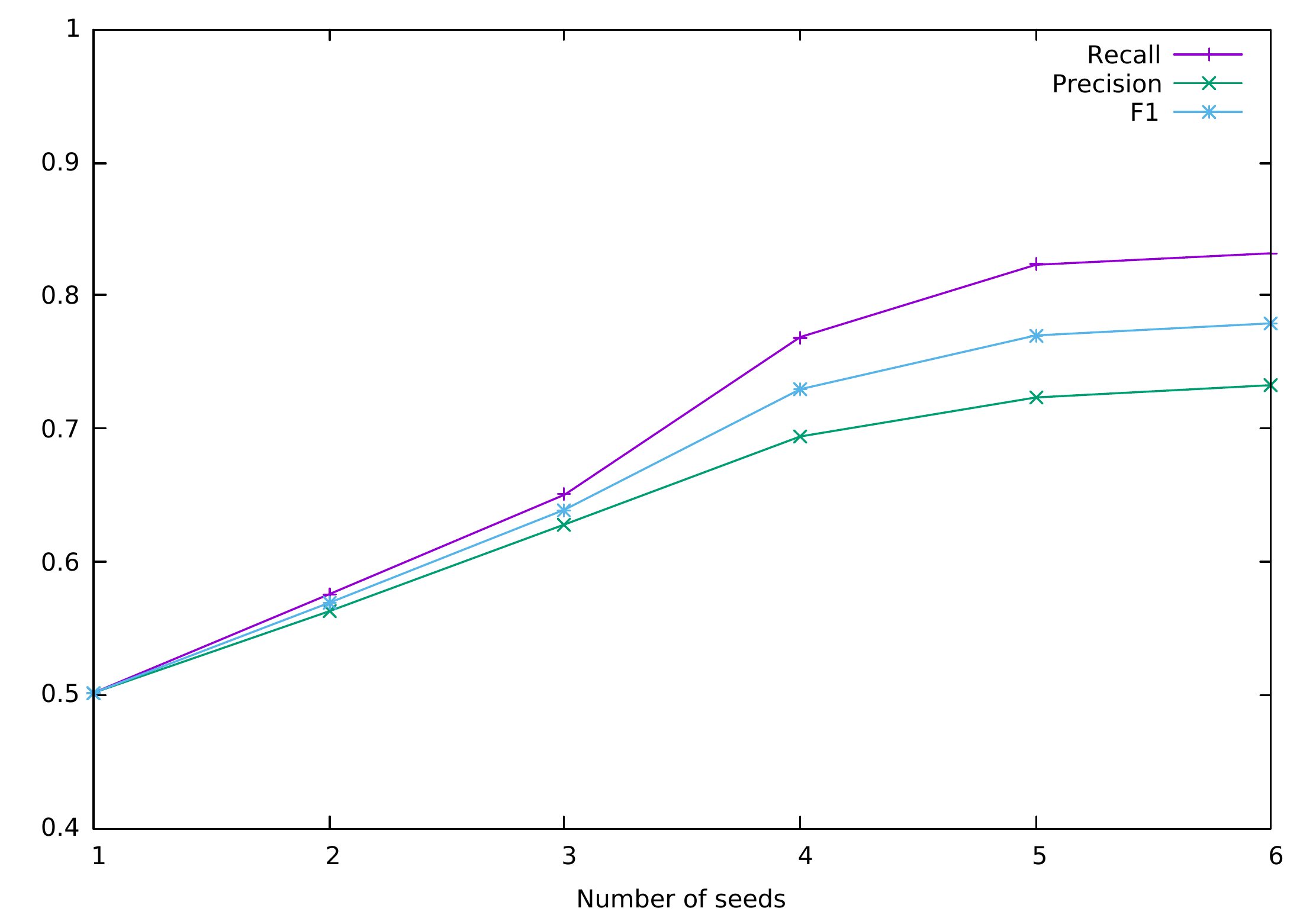}
    \caption{Recall, precision and F1 including the ``rest'' category}
    \label{fig:fscore_rest}
\end{figure}

Thus, within the extracted terminology, we have manually labeled about 100 entries
as neither drugs nor effects, and we have used them as candidates. This has been done to evaluate the effectiveness of using the parameter $\tratio$ to  avoid classifying these terms as either substances or effects. Performance-wise, this resulted in few more false positives given by terms erroneously assigned to the substance and effect classes, when instead these 100 candidates should ideally all be discarded. The results are in Table~\ref{tab:result3classes} and Figure~\ref{fig:fscore_rest}. We can observe that, when we include in the evaluation also those data that are neither substances nor effects, with no training data other than the original seeds, and operating only on the thresholds, the precision drops significantly. 

\begin{table}[ht]
    \centering
    \begin{tabular}{|c|c|c|c|}
        \hline
        \bf{\# of seeds} & \bf{Recall} & \bf{Precision} & \bf{F1}\\ 
        \hline
        1 & 0.502 & 0.502 & 0.502\\
        2 & 0.576 & 0.563 & 0.569\\
        3 & 0.650 & 0.628 & 0.639\\
        4 & 0.769 & 0.694 & 0.730\\
        5 & 0.823 & 0.723 & 0.770\\
        6 & 0.832 & 0.733 & 0.779\\
        \hline
    \end{tabular}
    \caption{Classification results for substances and effects, including the ``rest'' category}
    \label{tab:result3classes}
\end{table}

\begin{figure}[!ht]
    \centering
    \includegraphics[width=0.7\textwidth]{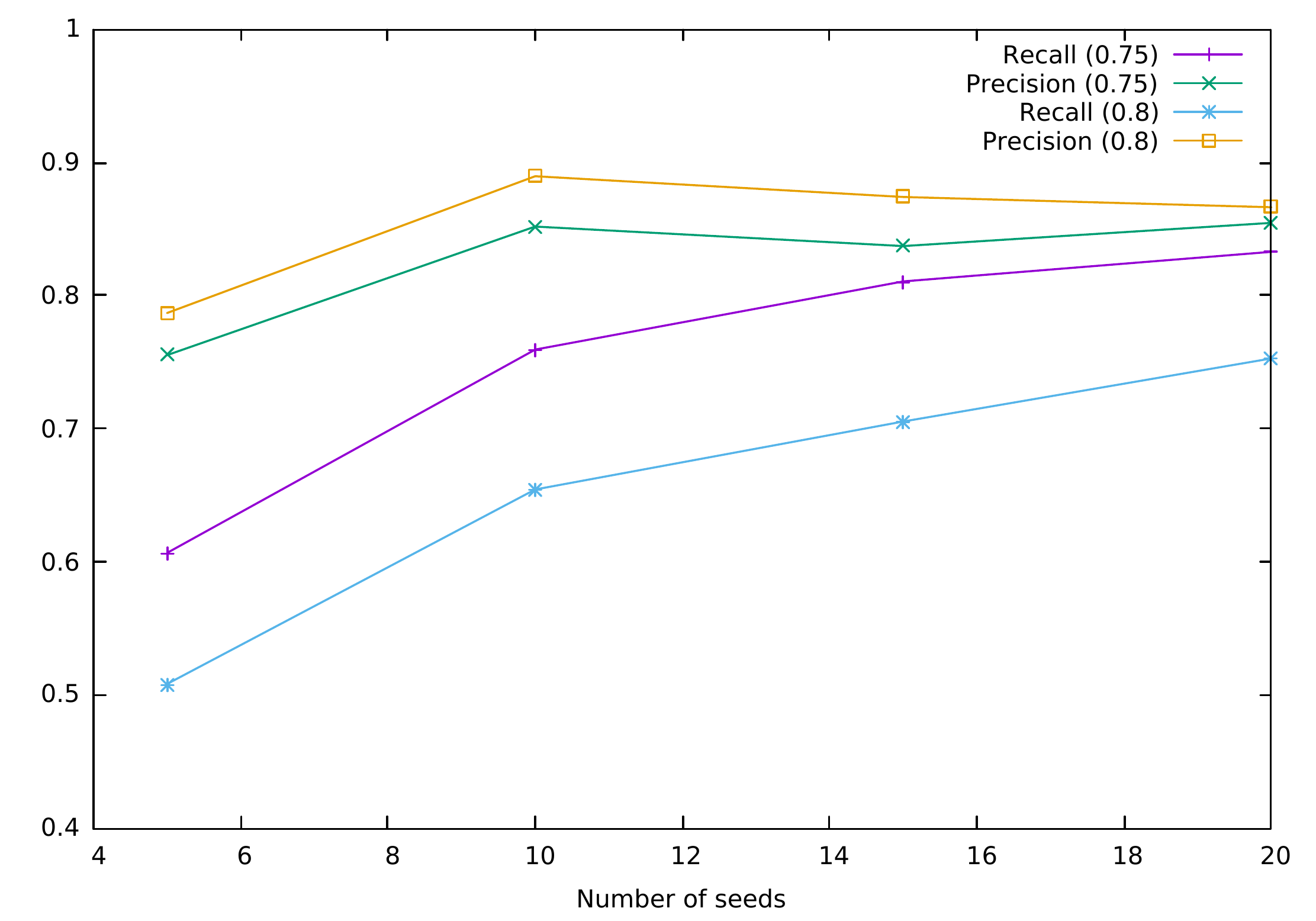}
    \caption{Precision and Recall with \tratio~set to 0.75 and 0.8 (incl. ``rest" category)}
    \label{fig:fscore_rest_table}
\end{figure}

\begin{figure}[!ht]
    \centering
    \includegraphics[width=0.7\textwidth]{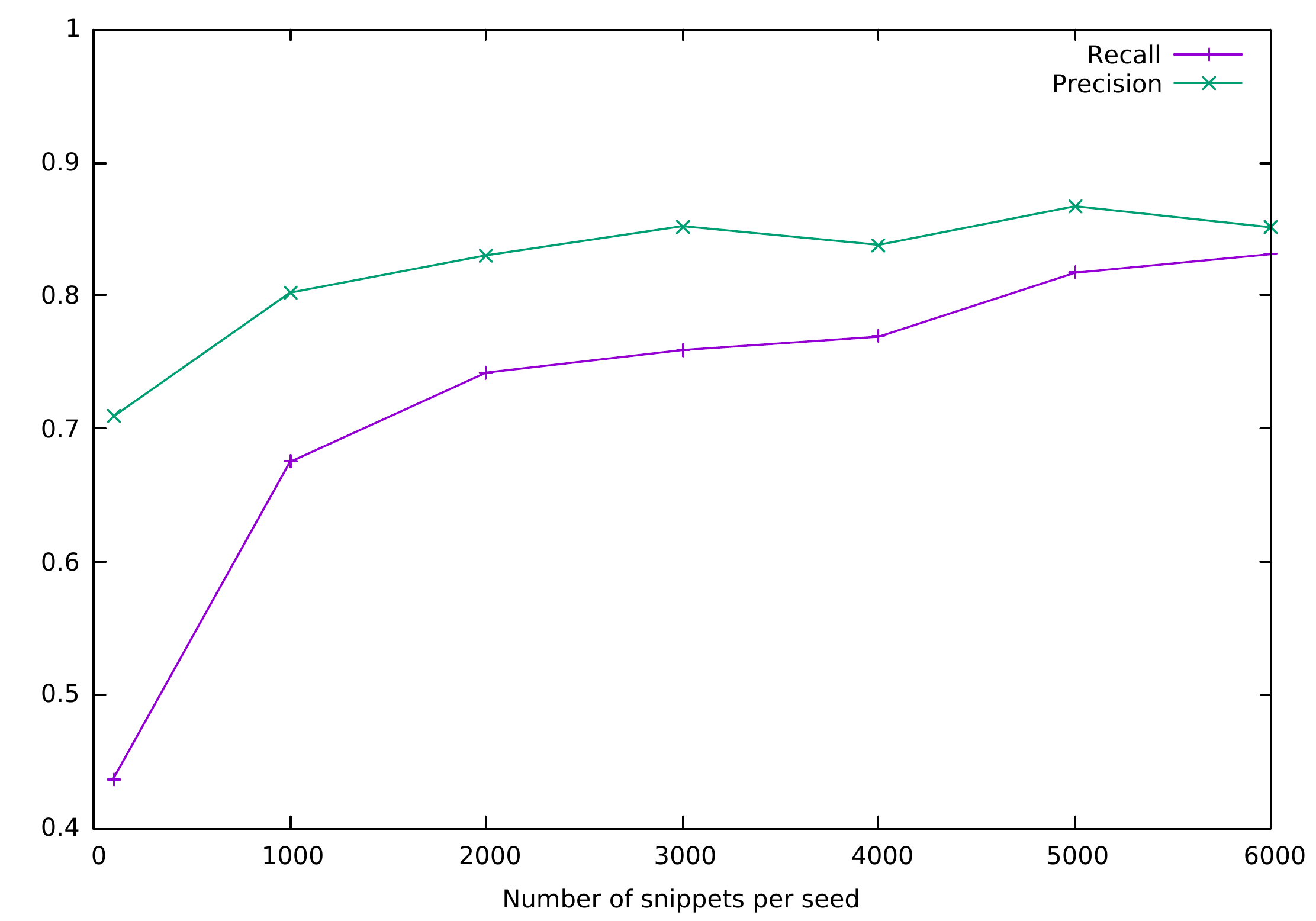}
    \caption{Recall and precision varying the number of contexts (snippets) per seed, 10 seeds used}
    \label{fig:fscore_snipseed}
\end{figure}

To achieve comparable performances, we have conducted experiments changing the number of seeds and \tratio~used to keep relevant terms.
The results are shown in Table~\ref{tab:resultsseedsthres} and Figure~\ref{fig:fscore_rest_table}. The higher the threshold, the higher the precision, while increasing the number of seeds improves the recall, which is to be expected: adding seeds ``teaches'' the system more about the variety of the data.
Moreover, recall augments when we increase the number of contexts per seed used to train the system (Table~\ref{tab:resultsniptrain} and Figure~\ref{fig:fscore_snipseed}).

\begin{table}[ht]
    \centering
    \begin{tabular}{|c|c|c|c|c|c|c|}
        \hline
        \bf{\# of seeds} & \bf{Recall 0.75} & \bf{Precision 0.75} & \bf{F1 0.75} & \bf{Recall 0.8} & \bf{Precision 0.8} & \bf{F1 0.8}\\ 
        \hline
        5 & 0.607 & 0.755 & 0.673 & 0.508 & 0.787 & 0.618\\
        10 & 0.759 & 0.852 & 0.803 & 0.654 & 0.889 & 0.754\\
        15 & 0.811 & 0.837 & 0.824 & 0.705 & 0.874 & 0.781\\
        20 & 0.833 & 0.854 & 0.843 & 0.753 & 0.866 & 0.805\\      
        \hline
    \end{tabular}
    \caption{Precision, Recall and F1 with \tratio~set to 0.75 and 0.8 (incl. ``rest'' category)}
    \label{tab:resultsseedsthres}
\end{table}

\begin{figure}[!ht]
    \centering
    \includegraphics[width=0.7\textwidth]{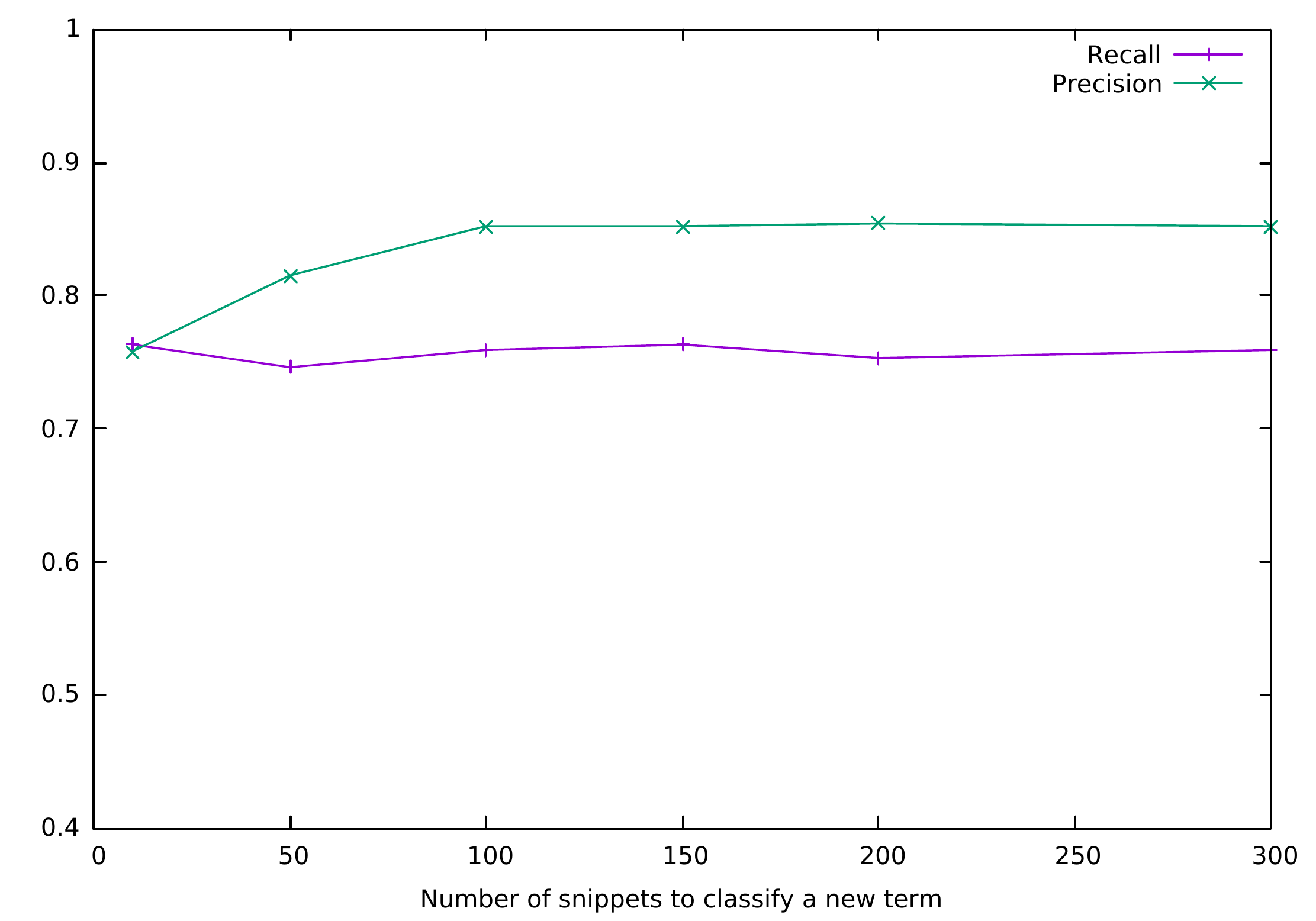}
    \caption{Recall and precision varying the number of contexts (snippets) per new term, 10 seeds used}
    \label{fig:fscore_sniptag}
\end{figure}

\begin{table}[!htb]
\begin{minipage}{0.5\linewidth}
    \centering
    \begin{tabular}{|c|c|c|c|}
        \hline
        \bf{\# of}  & \bf{Recall} & \bf{Precision} & \bf{F1}\\ 
        \bf{contexts} & & & \\
        \hline
        100 & 0.437 & 0.709 & 0.541\\
        1000 & 0.675 & 0.802 & 0.733\\
        2000 & 0.742 & 0.830 & 0.784\\
        3000 & 0.759 & 0.852 & 0.803\\
        4000 & 0.769 & 0.838 & 0.802\\
        5000 & 0.817 & 0.867 & 0.841\\
        6000 & 0.831 & 0.851 & 0.840\\      
        \hline
    \end{tabular}
    \parbox{5cm}{\caption{Results varying the number of contexts per seed\label{tab:resultsniptrain}}
    }
\end{minipage}%
\begin{minipage}{0.5\linewidth}
    \centering
    \begin{tabular}{|c|c|c|c|}
        \hline
        \bf{\# of } & \bf{Recall} & \bf{Precision} & \bf{F1}\\ 
        \bf{contexts} & & & \\
        \hline
        10 & 0.763 & 0.758 & 0.760\\
        50 & 0.746 & 0.815 & 0.779\\
        100 & 0.759 & 0.852 & 0.803\\
        150 & 0.763 & 0.852 & 0.805\\
        200 & 0.753 & 0.854 & 0.800\\
        300 & 0.759 & 0.852 & 0.803\\
        \hline
    \end{tabular}
    \parbox{5cm}{\caption{Results varying the number of contexts per new term\label{tab:resultsniptag}}}
    
    \end{minipage}
\end{table}



It is  worth noting that increasing the number of contexts used to classify a new term seems to have no effect after few contexts, as shown in Table~\ref{tab:resultsniptag} and Figure~\ref{fig:fscore_sniptag}). This indirectly conveys an information on the variety of contexts present on the investigated datasets.

Interestingly, the automated drug detection reported 1846 drugs in Bluelight and 1857 in DrugsForum, with 1520 drugs in common between the two forums. Moreover, some drugs appear exclusively in one of the two forums, like the \textit{triptorelin}, \textit{candesartan} and \textit{thiorphan} in Bluelight and the \textit{lymecycline}, \textit{boceprevir} and \textit{imipenem} in Drugsforum, although the majority is shared. 

Finally, upon training the system with the seeds, for every post it is possible to link the drugs to their effects. 
An example of links is in Table~\ref{tab:effects}.

\begin{table}
    \centering
    \begin{tabular}{|c|c|}
        \hline
        \bf{Drug} & \bf{Effects} \\
        \hline
        heroin &    anxiety, euphoria \\
        cocaine	&	euphoria, anxiety, comedown, paranoia, psychosis \\
        ketamine &	euphoria, anxiety, visuals,	comedown, hallucinations, nausea \\
        methadone &	anxiety, euphoria \\
        codeine	& euphoria,	anxiety, nausea	\\
        morphine & euphoria, anxiety, analgesic, nausea \\
        amphetamine	& euphoria,	anxiety, comedown, psychosis, visuals \\
        oxycodone &	euphoria, anxiety \\
        methamphetamine	& euphoria,	anxiety, psychosis,	comedown, paranoia \\
        dopamine & euphoria, anxiety, comedown,	psychosis \\
        \hline
    \end{tabular}
    \caption{Main effects of the most discussed drugs on Bluelight}
    \label{tab:effects}
\end{table}
\section{Related work}
\label{sec:relwork}

%
Recently, Academia has started mining online communities, to seek for comments on drugs and drugs reactions~\cite{Yang2014}. Indeed, forums and social networks offer spontaneous information, with abundance of data about experiences, doses, assumption methods~\cite{Davey2012,Delvigna2016}. Authors in~\cite{Nikfarjam2015} realized ADRMine, a tool for adverse drugs reaction detection. The tool relies on advanced machine learning algorithms and semantic features based on word clusters - generated from pre-trained word representation vectors using deep learning techniques. 
Also, intelligence analysis has been applied to social media to detect new outbreaking trends in drug markets, 
%
as in~\cite{Watters2012}. 
A raising phenomenon connected to the consumption of psychoactive substances is the adoption of nonmedical use of prescription drugs~\cite{Mackey2013}, such as sedatives, opioids, and stimulants. Even these drugs are often traded and advertised online by fake pharmacies~\cite{Katsuki2015,Freifeld2014}.
The amount of data available nowadays has made automated text analysis veer towards more machine learning-based approaches. Because complex tasks might require many training examples, however, there is a vivid study on unsupervised and semi-supervised approaches.
Our task encompasses identifying names in text, something often associated with named-entity extraction. Unsupervised methods such as~\cite{Smith2005} use unlabeled data contrasted with other data  assumed irrelevant - to use as negative examples - in order to build a classification model. Instead, we use seeds, a small set of examples, because the writers on forums often attempt not to mention drugs explicitly, resorting to paraphrases or nicknames, making a purely contrastive approach difficult to apply. 
Also, multi-level bootstrapping proved to be a valid improvement in information extraction~\cite{Riloff1999}; this techniques feature an iterative process to gradually enlarge and refine a dictionary of common terms. Our approach, instead, splits the problem of finding candidate terms and classifying them in two separate subproblems, the second of which is fed with a small number of annotated examples, i.e., the seeds. 
Co-training is a common technique~\cite{Blum1998} to evaluate whether to use an unlabeled piece of data as a training example: the idea is building different classifiers, and use the label assigned by one as a training example for another. In our case, we instead leverage the redundancy among the data, to ensure candidate examples are selected with a high degree of confidence.
Relation extraction is an even more complex task which seeks for the relationships among the entities.
This is relevant here, because substances can only be identified basing on their role in the sentence (since common names are often used to refer to them).  Work in~\cite{Rosenfeld2007} proposes a method  based on corpus statistics that requires no human supervision and no additional corpus resources beyond the corpus used for relation extraction. Our approach does not explicitly address relation extraction , but it exploits the redundancy of a substance (or effect) being often associated with other entities to identify them. KnowItAll~\cite{Etzioni2005} is a tool for unsupervised named entity extraction with improved recall, thanks to the pattern learning, the subclass extraction and the list extraction features that still includes bootstrapping to learn domain independent extraction patterns. 
For us, common mention patterns are also strong indicators of the substance or effect class; however, we do not use patterns to extract, but only, implicitly, for classification purposes. Furthermore, \cite{Carlson2010} pursues the thesis that much greater accuracy can be achieved by further constraining the learning task, by coupling the semi-supervised training of many extractors for different categories and relations; we use a single multiclass classifier to achieve the same goal. Under the assumption that the number of labeled data points is extremely small and the two classes are highly unbalanced, the authors of~\cite{Xie2011} realized a stochastic semi-supervised learning approach that was used in the 2009-2010 Active Learning Challenge. While the task is similar, our approach is different, because we do not need to use unlabeled data as negative examples.
The framework proposed in~\cite{Chang2007} suggests to use domain knowledge, such as dictionaries and ontologies, as a way to guide semi-supervised learning, so as to inject knowledge into the learning process. We have not relied on rare expert knowledge for our task, arguing that a few labeled seeds are easier to produce than dictionaries or other forms of expert knowledge representations. A mixed case of learning extraction patterns, relation extraction and injecting expert knowledge is in~\cite{Bellandi2010}, which also shows the challenge of evaluating a technique when few labeled examples are available.
As shown above, the problem of building a model with a limited set of information, but with a large enough amount of data, has been tackled by various angles. Our main staples were: a) the availability of a large set of unlabeled data, and b) the availability of a small set of labeled substance and effect names.



\section{Conclusions}
\label{sec:conc}
We have automatically identified and classified substances and effects from posts of drugs forums, making use of a semi-supervised text mining approach. Human intervention is required for the creation of a small training set, but the algorithm is able to automatically discover substances and effects with such a very few initial information. We believe our proposal will help sensitizing drug consumers about the risks of their choices and will contrast the diffusion of NPS, which spread on the online market at an impressive high rate. 
\vspace{-2mm} 
\section{Acknowledgements}
This publication arises from the project “CASSANDRA, (Computer Assisted Solutions for Studying the Availability aNd Distribution of novel psychoActive substances)" which has received funding from the European Union under the ISEC programme.\\
Prevention of and fight against crime [JUST2013/ISEC/DRUGS/AG/6414].

\bibliographystyle{splncs03.bst}
\bibliography{bibliography}
\end{document}